%% file: root.tex
\documentclass[letterpaper, 10 pt, conference]{ieeeconf}  

\IEEEoverridecommandlockouts                              

\usepackage{graphics} 
\usepackage{epsfig} 
\usepackage{times} 
\usepackage{amsmath} 
\usepackage{amssymb}  
\usepackage{float}
\usepackage{booktabs}
\usepackage{caption}
\usepackage{lipsum}
\usepackage{graphicx}
\usepackage{multirow}
\usepackage{xcolor} 
\usepackage{subfigure}
\usepackage{amsmath}
\usepackage{amssymb}
\usepackage{xspace}
\usepackage{threeparttable}

\title{\LARGE \bf DexRepNet: Learning Dexterous Robotic Grasping Network with Geometric and Spatial Hand-Object Representations}

\author{Qingtao Liu$^{1*}$, Yu Cui$^{1*}$, Qi Ye$^{1\dag}$, Zhengnan Sun$^{1}$, Haoming Li$^{1}$, Gaofeng Li$^{1}$, Lin Shao$^{2}$, and Jiming Chen$^{1}$ 
}

\begin{document}
\include{lib}

\newcommand{\yq}[1]{\textcolor{red}{\textbf{yq: }\xspace#1}\xspace}
\newcommand{\rep}{DexRep\xspace}
\newcommand{\method}{DexRepNet\xspace}
\newcommand{\occ}{f_o}
\newcommand{\surf}{f_s}
\newcommand{\loc}{f_l}
\newcommand{\policy}{\pi_\theta}
\newcommand{\hstate}{f_h}
\newcommand\blfootnote[1]{%
\begingroup 
\renewcommand\thefootnote{}\footnote{#1}%
\addtocounter{footnote}{-1}%
\endgroup 
}




\twocolumn[{%
\renewcommand\twocolumn[1][]{#1}%
\maketitle
\begin{center}
    \captionsetup{type=figure}
    \includegraphics[trim=0cm 8.7cm 0cm 1.2cm, clip, width=\textwidth]{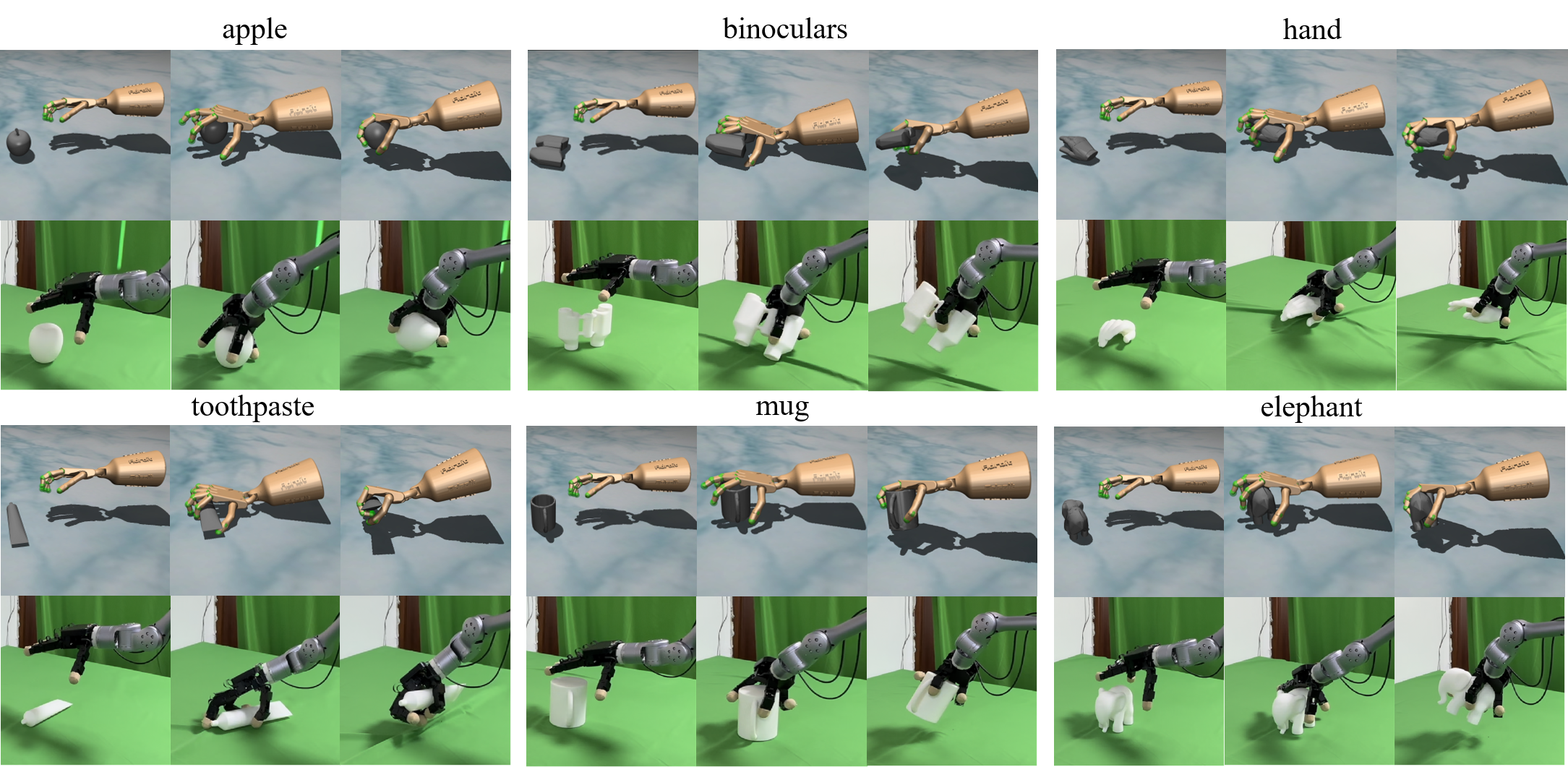}
    \captionof{figure}{Examples of our method evaluated in simulations with Adriot Hand of five fingers and in the real world with Allegro Hand of four fingers. Objects from left to right: apple, binoculars, hand.}
    \label{fig:teaser}
\end{center}%
}]

\thispagestyle{empty}
\pagestyle{empty}

\blfootnote{$^{1}$College of Control Science and Engineering, Zhejiang University, Hangzhou, 310027, China}
\blfootnote{$^{2}$Department of Computer Science, National University of Singapore, 15 Computing Dr, 117418, Singapore }
\blfootnote{$^{\dag}$ Corresponding Author Qi Ye (qi.ye@zju.edu.cn). Qi Ye is with the College of Control Science and Engineering and the State Key Laboratory of Industrial Control Technology, Zhejiang University, and also with the Key Lab of CS\&AUS of Zhejiang Province.}
\blfootnote{$^{*}$Equal contribution }
 
\input{text/abstract.tex}
\input{text/introduction.tex}

\input{text/method.tex}

\input{text/result.tex}
\input{text/conclusion.tex}


\bibliographystyle{IEEEtran}
\bibliography{mybibfile}

\end{document}

%% file: lib.tex
\newcommand{\reflabel}{dummy} 


\newcommand{\seclabel}[1]{\label{sec:\reflabel-#1}}
\newcommand{\secref}[2][\reflabel]{Section~\ref{sec:#1-#2}}
\newcommand{\Secref}[2][\reflabel]{Section~\ref{sec:#1-#2}}
\newcommand{\secrefs}[3][\reflabel]{Sections~\ref{sec:#1-#2} and~\ref{sec:#1-#3}}

\newcommand{\eqlabel}[1]{\label{eq:\reflabel-#1}}
\renewcommand{\eqref}[2][\reflabel]{(\ref{eq:#1-#2})}
\newcommand{\Eqref}[2][\reflabel]{(\ref{eq:#1-#2})}
\newcommand{\eqrefs}[3][\reflabel]{(\ref{eq:#1-#2}) and~(\ref{eq:#1-#3})}

\newcommand{\figlabel}[2][\reflabel]{\label{fig:#1-#2}}
\newcommand{\figref}[2][\reflabel]{Fig.~\ref{fig:#1-#2}}
\newcommand{\Figref}[2][\reflabel]{Fig.~\ref{fig:#1-#2}}
\newcommand{\figsref}[3][\reflabel]{Figs.~\ref{fig:#1-#2} and~\ref{fig:#1-#3}}
\newcommand{\Figsref}[3][\reflabel]{Figs.~\ref{fig:#1-#2} and~\ref{fig:#1-#3}}

\newcommand{\tablelabel}[2][\reflabel]{\label{table:#1-#2}}
\newcommand{\tableref}[2][\reflabel]{Table~\ref{table:#1-#2}}
\newcommand{\Tableref}[2][\reflabel]{Table~\ref{table:#1-#2}}
\newcommand{\etal}{et al.}
\newcommand{\eg}{e.g.}
\newcommand{\ie}{i.e. }
\newcommand{\etc}{etc. }

\def\bfmu{\mbox{\boldmath$\mu$}}
\def\bftau{\mbox{\boldmath$\tau$}}
\def\bftheta{\mbox{\boldmath$\theta$}}
\def\bfdelta{\mbox{\boldmath$\delta$}}
\def\bfphi{\mbox{\boldmath$\phi$}}
\def\bfpsi{\mbox{\boldmath$\psi$}}
\def\bfeta{\mbox{\boldmath$\eta$}}
\def\bfnabla{\mbox{\boldmath$\nabla$}}
\def\bfGamma{\mbox{\boldmath$\Gamma$}}

%
%


\newcommand{\R}{\mathbb{R}}

\newcommand{\be}{\begin{equation}}
\newcommand{\ee}{\end{equation}}

%% file: text/abstract.tex
\begin{abstract}
Robotic dexterous grasping is a challenging problem due to the high degree of freedom (DoF) and complex contacts of multi-fingered robotic hands. Existing deep reinforcement learning (DRL) based methods leverage human demonstrations to reduce sample complexity due to the high dimensional action space with dexterous grasping. However, less attention has been paid to hand-object interaction representations for high-level generalization. In this paper, we propose a novel geometric and spatial hand-object interaction representation, named \rep, to capture object surface features and the spatial relations between hands and objects during grasping.  \rep comprises Occupancy Feature for rough shapes within sensing range by moving hands, Surface Feature for changing hand-object surface distances, and Local-Geo Feature for local geometric surface features most related to potential contacts. 
Based on the new representation, we propose a dexterous deep reinforcement learning method \method to learn a generalizable grasping policy. 
Experimental results show that our method outperforms baselines using existing representations for robotic grasping dramatically both in grasp success rate and convergence speed. It achieves a 93\% grasping success rate on seen objects and higher than 80\% grasping success rates on diverse objects of unseen categories in both simulation and real-world experiments.
\end{abstract}

%% file: text/introduction.tex
\section{INTRODUCTION}
 \begin{figure*}[htbp]
     \centering
     \includegraphics[width=\textwidth]{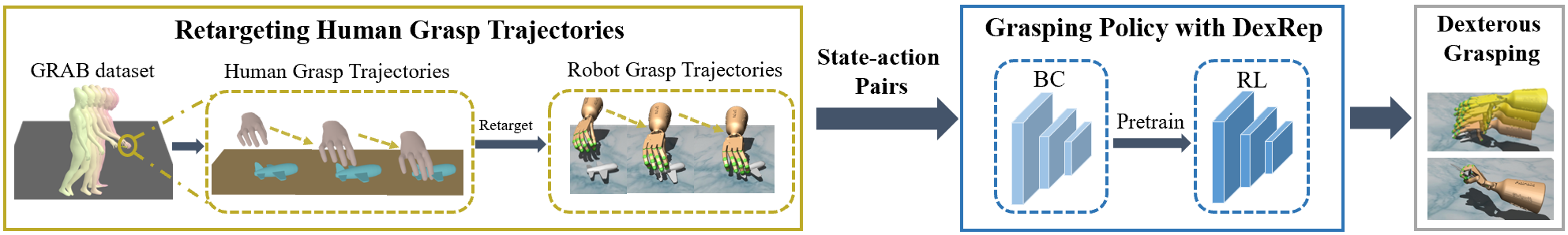}
     \caption{\textbf{The pipeline of DexRepNet.} We first retarget human grasp demonstrations to collect robot grasp trajectories (yellow), then we leverage the state-action pairs for grasp policy learning: behavior cloning for initialization and reinforcement learning for fine-tuning (blue). Finally, the trained policy can generate dexterous grasping with generalization (gray).
     }
     \label{fig:overview}
     \vskip -0.5cm
 \end{figure*}

Robotic dexterous grasping is a crucial skill to enable robots to perform complex tasks and achieve human-like capabilities in unstructured environments.
Despite the promising potential, learning dexterous grasping skills with multi-fingered hands is challenging due to its high DoF and complex contacts. Dexterous hands are typically equipped with many articulated joints, leading to more than 20 DoFs and formidable action space. Also, adapting the grasping skill to new objects requires learning methods to achieve good generalization to complex interaction contacts between multi-fingered hands and diverse objects.

Deep reinforcement learning (DRL) has shown great potential to solve the problem of dexterous grasping with its outstanding performance in sequential decision-making problems. DAPG~\cite{DAPG:rajeswaran2017learning} exploits human demonstrations collected in virtual reality to improve sample efficiency of DRL in a high-dimensional 30 DoFs hand space. Following the same human demonstration strategy, ILAD~\cite{ILAD:wu2022learning}, GRAFF~\cite{GRAFF:mandikal2021learning}, and DexVIP~\cite{mandikal2022dexvip} extends the demonstrations into larger scale trajectories or grasping poses collected from human interaction with real daily objects. GRAFF~\cite{GRAFF:mandikal2021learning} leverages contacts from hand-object interactions for agents to learn to approach objects more effectively.  DexVIP~\cite{mandikal2022dexvip} learns human pose priors from videos and imposes these priors into DRL by incorporating auxiliary reward functions favoring robot poses similar to the human ones in videos. ILAD~\cite{ILAD:wu2022learning} trains a generator to synthesize grasping trajectories with large-scale demonstrations instead of using human demonstrations directly.

Significant progress has been made in improving dexterous grasping performance through the above demonstrations to reduce sample complexity in the high-dimensional space. However, less attention has been paid to hand-object interaction representations for high-level generalization with dexterous hands. For point cloud inputs, features such as position, distance, or/and a global feature vector extracted by PointNet~\cite{qi2017pointnet} describe the relationship between hands and objects~\cite{wei2022dvgg},~\cite{shao2020unigrasp},~\cite{ye2022CGF}; for input with RGB images, a global feature vector extracted from CNN like ResNet~\cite{He_2016_CVPR} is used to capture the relation~\cite{levine2018handeye}. Though able to capture the geometry of objects, global features typically have difficulty in generalizing to new data. Moreover, simple representations like distances or positions of the objects and hands fail to capture the interactions between the diverse object surfaces and complex hand articulations, which is important for dexterous grasping having a large number of potential contacts. Therefore even with large-scale demonstrations, these works are still limited in the generalization to unseen objects. The policy learned by ILAD~\cite{ILAD:wu2022learning} has demonstrated good performance for objects in the same category. The grasp success rates of GRAFF~\cite{GRAFF:mandikal2021learning} and DexVIP~\cite{mandikal2022dexvip} for unseen objects in the simulation are lower than 70\%.

Aiming at improving generalization for dexterous grasping, in this paper, we propose a novel compound geometric and spatial hand-object representation to capture object surface features and the spatial relations between hands and objects when hands approach objects, which we name as \rep.  The compound representation comprises three components to fully capture the dynamics during interactions and embrace generalization at the same time:  1) Occupancy Feature, representing the voxels occupied by an object when a volume of low resolution attaching to a robotic hand moves towards the object; 2) Surface Feature, representing the closest distance of points on an object surface to a fixed set of points on a hand surface and also the normals of the closest points on the object; 3) Local-Geo Feature: representing the local geometry feature of the closest points. Occupancy Feature captures global shape information seen by the approaching hands via a coarse occupancy volume instead of static point clouds carrying detailed geometry and Local-Geo Feature only captures local detailed geometry: the combination of the coarse global and finer local features fully captures object surface information and also ensures generalization to unseen objects resembling objects in the training set roughly or partially. Surface Feature and Local-Geo Feature capture the dynamics of the interaction and the geometry feature of the most related object surface area for each hand part in potential contacts. 

Based on the new representation, we propose a dexterous deep reinforcement learning method with human demonstrations to learn a generalizable grasping policy \method. Fig.\ref{fig:overview} shows the pipeline of our method. To obtain robotic demonstrations across different objects, we collect human grasp demonstrations from GRAB~\cite{taheri2020grab} and use motion retargeting technologies~\cite{handa2020dexpilot} to transfer them to robotic grasping demonstrations for behavior cloning (BC). \rep is utilized to extract the representation information of hand-object interaction and input it together with hand states to train the grasping policy.

In summary, our main contributions in this paper are:
\begin{itemize}
    \item We propose a novel geometric and spatial hand-object interaction representation consisting of Occupancy Feature, Surface Feature, and Local-Geo Feature for robotic dexterous grasping
    \item We propose a dexterous deep reinforcement learning method based on the novel representation to learn a generalizable grasping policy.
    \item Our proposed method outperforms baselines using existing representation dramatically for seen and unseen objects both in simulation and in the real world. Experiments also demonstrate its effectiveness on various multi-finger hands.
\end{itemize}

\section{related work}


\textbf{Dexterous Robotic Grasping.} 
There have been works in different directions attempting dexterous grasping.
Liang \etal ~\cite{liang2021multifingered} map low-DOF end-effectors to high-DOF ones to solve the high-DOF grasping problem. Shao~\cite{shao2020unigrasp} models the robotic hand and object and uses optimization methods to find force closure and contact points.  Wei \etal~\cite{wei2022dvgg} presents an efficient grasp generation network that takes a single-view point cloud reconstructed by a point completion module as input and predicts high-quality grasp configurations for unknown objects. Turpin \etal~\cite{turpin2022graspd}  and Liu \etal~\cite{liu2020deep} adopt the differentiable simulation method to optimize a path towards stable grasping. To facilitate exploration and reduce sampling complexity, most DRL approaches for dexterous grasping require the use of expert demonstrations~\cite{DAPG:rajeswaran2017learning, ILAD:wu2022learning}. ~\cite{DAPG:rajeswaran2017learning} presents a demo augmented policy gradient method, incorporating expert trajectories collected via teleoperate into RL. \cite{ILAD:wu2022learning} proposed a novel imitation learning, using expert demonstrations to accelerate the phase of RL training. In comparison, some works incorporate human grasping affordance (contact map and grasping pose)~\cite{GRAFF:mandikal2021learning,mandikal2022dexvip, Christen_2022_CVPR}. ~\cite{GRAFF:mandikal2021learning} inject a visual affordance prior to deep RL grasping policies for functional grasping. Its following-up work~\cite{mandikal2022dexvip} learns dexterous grasping from in-the-wild hand-object interaction videos. Both works 
 aim to reduce or avoid the use of costly expert demonstrations. ~\cite{Christen_2022_CVPR} proposed dynamic grasp generation, which uses static grasp references to generate grasp motions. In this paper, we follow the human demonstration strategy and adopt DAPG ~\cite{DAPG:rajeswaran2017learning} for learning.

\textbf{Hand-Object Representation for Grasping.} DRL methods emphasize self-exploration in the interaction with the environment, therefore, a detailed description of the interaction is crucial. For object description, there are RGB~\cite{levine2018handeye}, depth~\cite{liu2020deep}, point cloud~\cite{wei2022dvgg, li2022contact2grasp}, mesh~\cite{varley2017shape}, and their fusion~\cite{GRAFF:mandikal2021learning},~\cite{mandikal2022dexvip},~\cite{cao2021suctionnet}. The description for hand is mainly the hand state~\cite{GRAFF:mandikal2021learning},~\cite{mandikal2022dexvip, yuan2017bighand2}. Unigrasp~\cite{shao2020unigrasp} introduced the URDF model of the hand so that the work can be applied to different multi-fingered dexterous hands without retraining the model. The hand-object relationship mainly uses more global information, such as the distance between hand and object~\cite{GRAFF:mandikal2021learning},~\cite{mandikal2022dexvip}, motor signals~\cite{joshi2020robotic}, etc. Though these descriptions work well in low-DoF grippers, their effectiveness in acquiring advanced dexterity for grasping skills is constrained due to their inability to fully capture the intricate interactions between various object surfaces and the intricate movements of the hand. ManipNet~\cite{zhang2021manipnet} introduces hand-object spatial representations to predict object manipulation trajectories given wrist trajectories by supervised learning for animations. Inspired by it, we propose to design more comprehensive hand-object interaction representations to empower reinforcement learning for dexterous grasping skills.


%% file: text/method.tex
\section{Method}

Fig.\ref{fig:overview} shows the pipeline of our method. We build our method upon DAPG~\cite{DAPG:rajeswaran2017learning}, which is a two-stage learning strategy of behavior cloning and DRL for a grasping policy. The state-action pairs for our policy learning are prepared by retargeting the human grasp trajectories to a dexterous robotic hand. Therefore, our method consists of three stages: retargeting, behavior cloning, and reinforcement learning, which is elaborated in \secref{dexrepnet}. Our proposed state representation of the hand-object interaction for the policy is detailed in \secref{dexrep}.

We model the problem of dexterous grasping as Markov Decision Process (MDP), which is defined by a tuple: $(\mathcal{S}, \mathcal{A}, \mathcal{T}, \mathcal{R}, \gamma)$. $\mathcal{S}$ and $\mathcal{A}$ are the state and action space.  The policy $\pi_\theta: \mathcal{S} \rightarrow \mathcal{A}$ maps the state space to the action space. $\mathcal{T}:\mathcal{S} \times \mathcal{A} \rightarrow \mathcal{S}$ is the transition dynamic. $\mathcal{R}:\mathcal{S} \times \mathcal{A} \rightarrow \mathbb{R} $ is the reward function and $\gamma\in\left ( 0, 1 \right ] $ is the discount factor. 
We use the Adroit platform~\cite{kumar2013adroithand} as our manipulator which consists of a 24-Dof hand attached to a 6-Dof arm, thus the action space is the continuous motor command of 30 actuators.

\subsection{Grasping policy with DexRep} \seclabel{dexrep}
\begin{figure}[htbp]
     \centering
     \includegraphics[width=\linewidth]{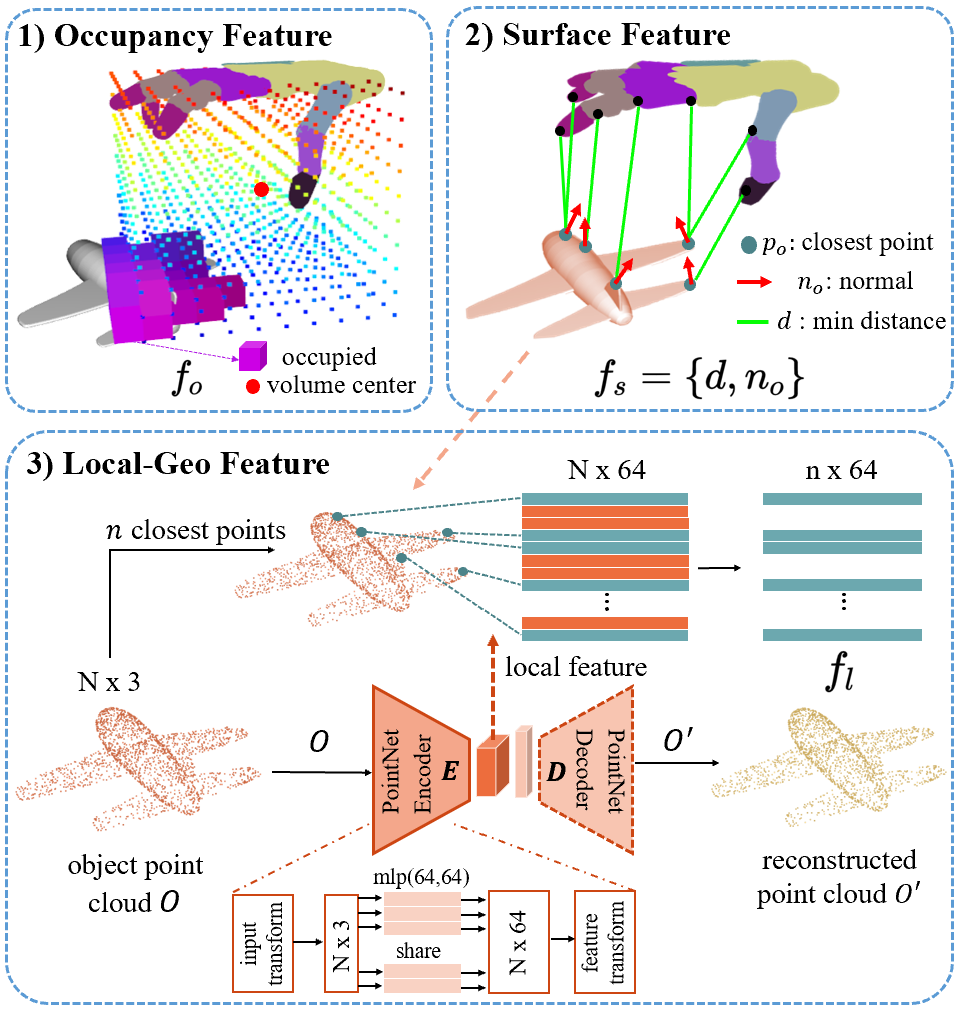}
     \caption{\textbf{DexRep.} Representation for dexterous grasping, describing both geometric and spatial hand-object representation. DexRep consists of three components: 1) Occupancy Feature 2) Surface Feature and 3) Local-Geo Feature.
     }
     \label{fig:DexRep}
     \vskip -0.5cm
 \end{figure}

In this part, our proposed \rep for dexterous grasping considering both representation power and the generalization capability are first introduced, followed by the design of the policy $\pi_{\theta}$ with \rep.

\subsubsection{Occupancy Feature $\occ$}

Occupancy Feature represents the voxels occupancy states when a grid volume attached to a robotic hand approaches the object. It captures global shape information via a coarse occupancy volume instead of point clouds carrying detailed geometry for high-level generalization.  Specifically, the occupancy volume is a $20\times 20\times 20cm^3$ volume with each voxel of $2\times 2 \times 2 cm^3$. We fix the volume center at a point on the inside part of the palm by an offset from the root joint of the index finger (illustrated at the top left of Fig.\ref{fig:DexRep} ) as the inner palm space is our interest region for grasping not the space above the back of the hand. We set the offset perpendicular to the palm plane. The occupancy volume moves and rotates with the root joint. We define Occupancy Feature as a vector $\occ \in \mathbb{R}^{1000}$ and for the $i^{th}$ item, it denotes the corresponding $i^{th}$ voxel occupied or not:

\be
    f_o^{i}=\left\{\begin{matrix}
  1&,occupied \\
  0&,otherwise
\end{matrix}\right.
\ee

The feature describes the rough shape of an object close to the hand. It helps the hand approach objects without collision. By increasing the resolution of the volume, we can perceive finer object surfaces. However, it not only increases the computational burden but also reduces the network's generalization ability to different objects. 

\subsubsection{Surface Feature $\surf$ }
When the robotic hand is near the object, the surface area around the hand becomes important cues for grasping. Therefore, we propose to represent this area by Surface Feature $\surf \in \mathbb{R}^{4n}$,  where $n$ is the number of fingertips and joints of a hand.  It is defined as: 
\be
f_s^{j}=\left\{\min \left(\left\|{p}_{j}-{p}_{o}\right\|, \sigma _{\max }\right), {n}_{o}\right\},
\ee
where $j$ denotes the $j^{th}$ point on the hand, ${p}_{o}$ is the closest point on the object surface to the$j^{th}$ point ${p}_{j}$ and  $\left\|{p}_{j}-{p}_{o}\right\|$ represents the distance. The distance has a maximum value of $\sigma _{\max }=20cm$. ${n}_{o}$ is the surface normal of the closet point. Fig.\ref{fig:DexRep} (top right) illustrates the definition. Surface Feature contains two aspects: (\romannumeral1) the distances to inform the hand where to touch; (\romannumeral2) the normal vectors to inform the hand what will be touched. The distance features provide the robotic hand with the local information of the object region closest to it. The local information can help the hand decide how to take action to touch and grasp the object.

\subsubsection{Local-Geo Feature $\loc$} 

Though the surface normals of the closest points above depicted the geometric feature of the surface points, more abundant geometric local features like curvatures, thickness, symmetry \etc which are important for grasping are not captured. Learning-based descriptors have been demonstrated to be able to represent comprehensive features for the inputs. Therefore, we propose to extract local geometric features for these object points by PointNet~\cite{qi2017pointnet}.

The extraction consists of two steps: feature learning and feature extraction. In the feature learning step, an auto-encoder taking a point cloud $O\in \mathbb R^{N\times3}$ as input is constructed in Fig.\ref{fig:DexRep} (bottom) to recover the point cloud $O^{\prime}$. 
 We train the auto-encoder using 42k objects in 55 categories in ShapeNet55~\cite{chang2015shapenet}. In the feature extraction step, we only keep the encoder part to acquire local features $f_l\in \mathbb R^{n\times64}$ for the $n$ closest points.  The two-stage feature extraction is adopted instead of learning the encoder jointly with a grasping policy $\pi_\theta$ since DRL only provides sparse rewards which are hard to learn the features for diverse objects.

\begin{figure}[htbp]
     \centering
     \includegraphics[width=\linewidth]{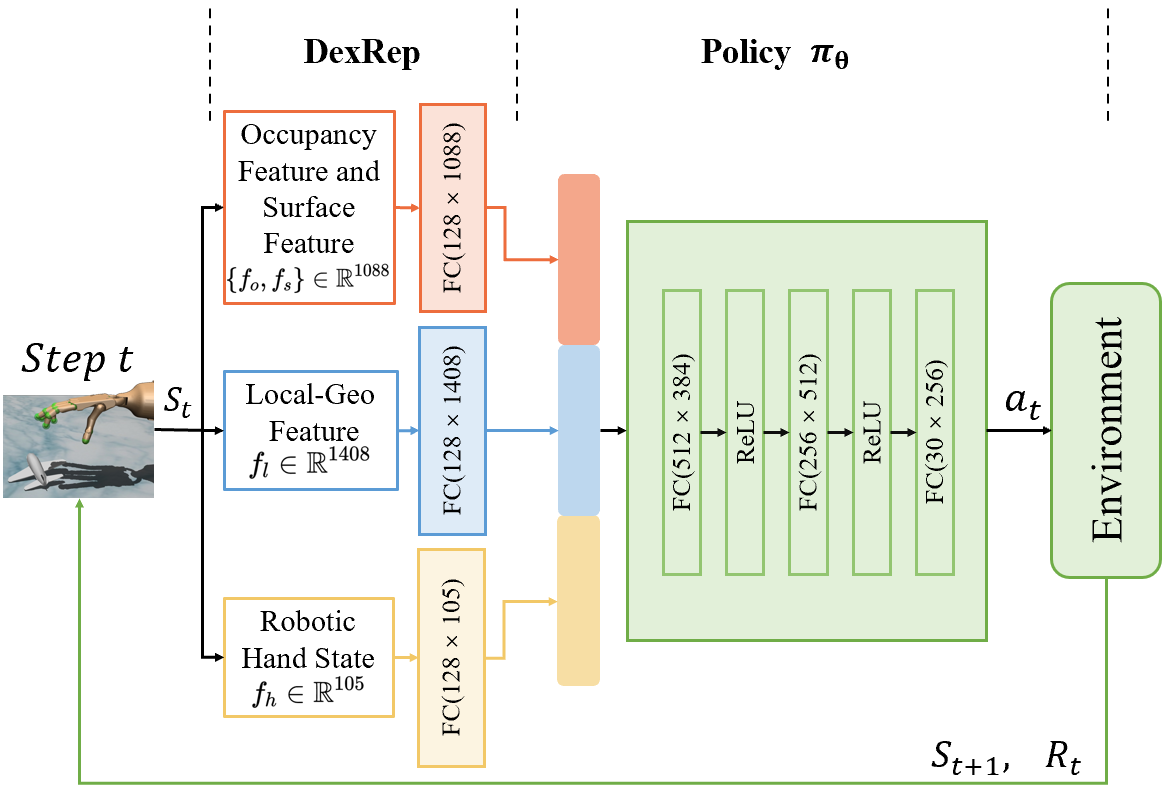}
     \caption{\textbf{Policy network architecture.} The policy maps DexRep and robotic hand state to actions.}
     \label{fig:policy}
 \end{figure}

 \subsubsection{Grasping policy $\policy$}
 
 The policy network architecture is illustrated in Fig.\ref{fig:policy}. In addition to DexRep mentioned above, we add the state of the robotic hand, which is represented as $f_{h}$, to the input of the policy network. $f_{h}$ consists of the joint angles of the Adroit hand, the positions of the joints and fingertips relative to the wrist, and the global translation and rotation of the hand root.   

Before fed into the policy network, the robotic hand state and DexRep are processed respectively by a single-layer fully-connected encoder with an output size of $128$ and an input size of $105$, $1088$ and $1408$ for $f_{h}$, $\left\{f_o, f_s\right\}$ and $f_{l}$. The weight of the fully-connected layer is fixed after behavior cloning. The policy network is an MLP with hidden layers $\left[ 512, 128\right]$, and ReLU activations between each layer.
\begin{figure*}[htbp]
    \centering
    \begin{minipage}{0.32\linewidth}
    \centering
    \includegraphics[width=\linewidth]{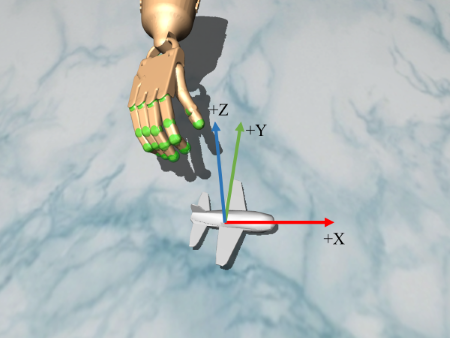}
    \end{minipage}
    \hfill
    \begin{minipage}{0.65\linewidth}
    \includegraphics[width=\linewidth]{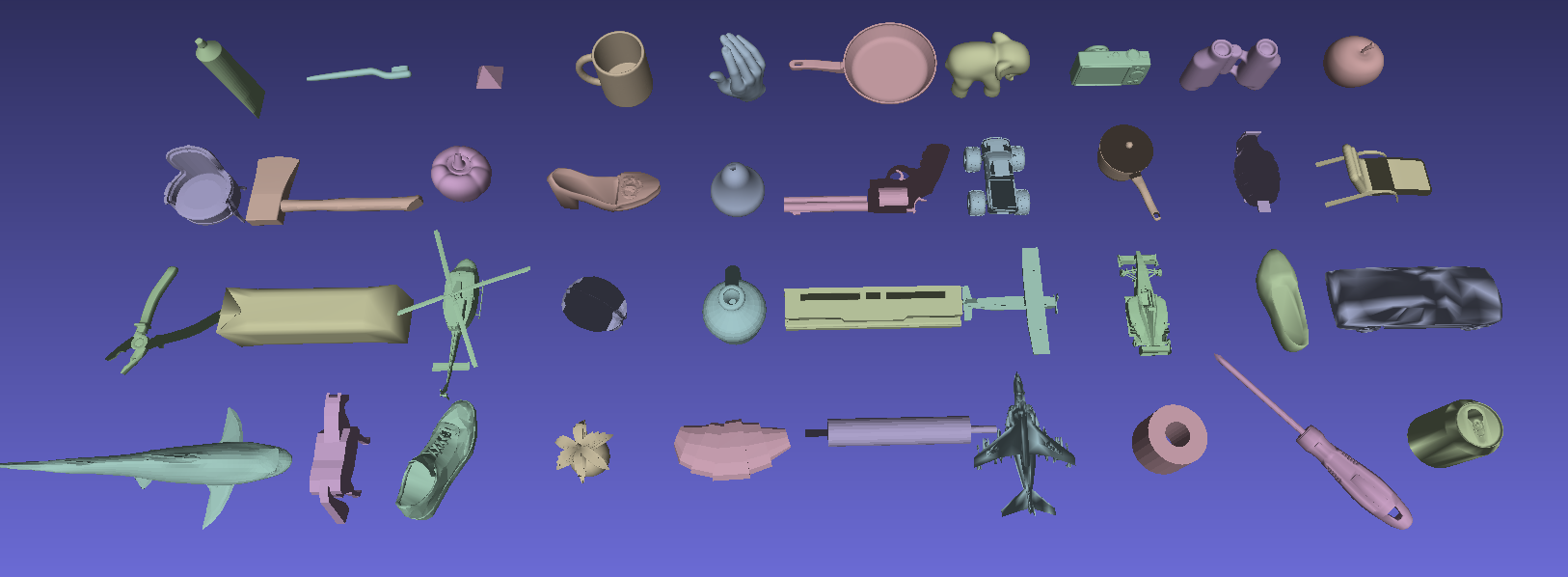}
    \end{minipage}
    \caption{The left is MuJoCo simulation environment and initial pose of the hand; The right is 40 novel objects used to evaluate our method. The objects on the first row is from GRAB~\cite{taheri2020grab} and the rest is selected randomly from 3DNet~\cite{wohlkinger20123dnet}.}
    \label{fig:expinit}
    \vskip -0.5cm
\end{figure*}
 \subsection{Learning DexRepNet} 
 \seclabel{dexrepnet}
In this section, we illustrate DexRepNet, the full learning pipeline for dexterous grasping: retargeting human grasp demonstrations, behavior cloning for policy initialization, and reinforcement learning for policy fine-tuning.

 \subsubsection{Retargeting Human Grasp Trajectories} 
 \label{retarget}
 
We first introduce the process of generating robotic grasp demonstrations. The trajectories of human demonstrations are from GRAB~\cite{taheri2020grab}, a dataset that contains a large number of whole-body pose sequences of 10 subjects interacting with 51 daily objects. 
We first extract hand-object interaction sequences with the right MANO hand~\cite{MANO:SIGGRAPHASIA:2017}. Then, we preserve every sequence from the frame where the distance between the hand and the object is $10cm$ to the frame where the object is lifted off the table about $4cm$. Lastly, we move the object to the coordinate without any rotation and translate the hand accordingly. 
We define human grasp demonstrations $\mathcal{D}_{\text {Human }}=\left\{d_{0},\ldots, d_{i}, \ldots, d_{I}\right\}$, where the $i^{th}$ demonstration 
$d_{i}=\left \{ \left ( g_{t}, q_{t}, o_{t}\right )_{t=0}^{T}\right\}$ 
is a collection of human hand global pose $g_{t}$, joint pose $q_{t}$ and object 6D pose $o_{t}$ of the $t^{th}$ step.


Given human demonstrations, our goal is to map them to a robotic hand. Following DexPilot~\cite{handa2020dexpilot}, we formulate the retargeting objective as a nonlinear optimization problem, with a cost function that makes the posture of the robotic hand consistent with that of the MANO hand~\cite{MANO:SIGGRAPHASIA:2017}, as well as makes the distance between the hand and the object consistent, ensuring the accuracy of the grasping position. Therefore, we define the retargeting objective similar to the function in~\cite{handa2020dexpilot} as:
\be 
\min_{q_{t}^{\mathbf{R}},M_g}\sum_{i=0}^{N}\left\|\mathbf{v}_{\mathbf{i}}^{\mathbf{R}}\left({M}_g , q_{t}^{\mathbf{R}}\right)-k_i\mathbf{v}_{\mathbf{i}}^{\mathbf{H}}\left(q_{t}^{\mathbf{H}}\right)\right\|^{2},
\eqlabel{retarget}
\ee
where $\mathbf{v}_{\mathbf{i}}^{\mathbf{R}}$ and $\mathbf{v}_{\mathbf{i}}^{\mathbf{H}}$ computed by the joint angles $q_{t}^{\mathbf{R}}$ and $q_{t}^{\mathbf{H}}$ through forward kinematics, respectively represent three kinds of key vectors (finger-to-finger vectors, finger-to-wrist vectors, finger-to-object vectors) of the robotic hand and human hand in the $t^{th}$ step, $k_i$ is the scale ratio of each vector between Adroit Hand~\cite{kumar2013adroithand} and MANO Hand~\cite{MANO:SIGGRAPHASIA:2017}. $M_g$ is the global translation and rotation of the arm attached to the robotic Hand. We find that by including the arm in the optimization process, the position error caused by the structural difference between the human hand and the robotic hand can be reduced. After optimization, we convert the optimized joint angles to actions in MuJoCo~\cite{todorov2012mujoco} and perform correlated sampling~\cite{chen2022dextransfer} on actions in case of dropping the object when lifting without sufficient grasping force. Finally, we execute the refined action sequences in simulation in order to collect demonstrations $\mathcal{D}$ composed of state-action pairs $\left(s,a\right)$, which are for behavior cloning in the next stage.

 \subsubsection{Pretraining with Behavior Cloning} 
 \seclabel{bc}

For high DoF robotic hands, the sample complexity is tremendous and RL from scratch may fail frequently. The idea of pretraining grasping policy with human demonstrations has worked successfully in prior works~\cite{DAPG:rajeswaran2017learning},~\cite{ILAD:wu2022learning}. Therefore, we follow the same strategy. 
With  $\mathcal{D}$, we pretrain the policy $\policy$ with the objective:
\begin{equation}
    L_{bc}=\frac{1}{|\mathcal{D}|}\sum _{(s,a)\in \mathcal{D}} \left\|\pi_{\theta }(s)-a\right\|^2
\end{equation}
where $|\mathcal{D}|$ is the number of state-action pairs in the dataset and $\theta$ is the parameter of the policy network to be optimized.  

 \subsubsection{RL fine tuning} 
 \seclabel{rl}

Though BC with large-scale human demonstrations can provide good policy initialization, the demonstration information is not fully used. DAPG~\cite{DAPG:rajeswaran2017learning} incorporates demonstrations into RL training to better assist the network in learning by adding an extra term to compute the gradient: 
\begin{equation}
    \begin{aligned}
    g_{\text {aug }}= & \sum_{(s, a) \in \rho_{\pi}} \nabla_{\theta} \ln \pi_{\theta}(a \mid s) A^{\pi_{\theta}}(s, a)+ \\
    & \sum_{(s, a) \in {D}} \nabla_{\theta} \ln \pi_{\theta}(a \mid s) \lambda_{0} \lambda_{1}^{k} \max _{(s, a) \in \rho_{\pi}} A^{\pi_{\theta}}(s, a),
    \end{aligned}
\end{equation}
where $\rho_{\pi}$ and $\mathcal{D}$ are the sampled dataset by the policy $\pi_{\theta}$ and the collected demonstrations respectively. $A^{\pi_{\theta}}$ is the advantage function, $\lambda_0=0.1$ and $\lambda_1=0.95$ are hyper-parameters, and $k$ is the iteration counter. In our experiments, we use GAE~\cite{GAE:schulman2015high} as the advantage function $A^{\pi_{\theta}}$ and set $\max _{(s, a) \in \rho_{\pi}} A^{\pi_{\theta}}(s, a)$=1.

For dexterous grasping, we design the reward function as follows:
\begin{equation}
    r=r_{d}+r_{g}+r_{t}.
\end{equation}

$r_{d}$ is the approaching reward, which incentives the hand to approach the object. In order to adapt more object geometries, we take advantage of the sum of the closet distance $d_{f2o}$ between all fingertips and the object surface, thus $r_{d}$ is defined as:
\begin{equation} \label{reward distance}
   r_{d}= \alpha \left ( \frac{1}{10\cdot d_{f2o}+0.25} -1  \right ) ,
\end{equation}
where $\alpha$ is the weight and set to 0.1.

$r_g$ is the grasping reward and is defined as:
\begin{equation} \label{reward grasping}
    r_g=\left\{\begin{matrix}
  1&,h>0.02 \\
  0&,0\le h\le 0.02
\end{matrix}\right.
\end{equation}
where $h$ is the lifting height of the object.

$r_t$ is designed to motivate the hand to move the lifting object to a target position. In our experiment, the target position $p_{tar}$ is $0.15m$ higher than the initial position $p_{obj}^0$. $r_t$ is defined as:
\begin{equation} \label{reward target}
    r_t=\left\{\begin{matrix}
  10&,\left \| p_{obj}^0-p_{tar} \right \|^2< 0.1   \\
  20&, \left \| p_{obj}^0-p_{tar} \right \|^2< 0.05 \\
  0&, otherwise
\end{matrix}\right.
\end{equation}
 
Eq.\ref{reward distance} guides the hand to touch the object, thereby facilitating the learning of a grasping policy. Eq.\ref{reward grasping} and Eq.\ref{reward target} guarantee that the learned policy can attain a stable grasping pose and reach the desired position.


%% file: text/result.tex
\section{experiment results}

In this section, we conduct a series of experiments to answer the following questions: 

\begin{enumerate}
    \item Can the proposed method grasp diverse objects and generalize well to unseen objects?    
    \item Can our method be applicable to various dexterous hands and keep as good performance as the five-fingered robotic hand?    
    \item Does our proposed feature work well with depth sensors with noise in the real world?
\end{enumerate}

Before these experiments, we first introduce the simulation and the object dataset in Section \ref{exp settings} and then we list a few baselines used to compare with our method in Section \ref{baselines}.
\begin{table*}[htbp]
      \centering
   \tabcolsep=2.0pt
   \small
  \caption{Success rate (\%) on 10 unseen objects of GRAB compared with baselines}
  \begin{threeparttable}
    \begin{tabular}{l|cccccccccc|c}
    \toprule
    \toprule
    Methods & mug   & camera & binoculars & apple & toothpaste & fryingpan & toothbrush & elephant & hand  & pyramidsmall & average \\
    \midrule
    Hand2obj\tnote{*} & 31.67  & 32.00  & 30.33  & 33.33  & 18.67  & 25.33  & 2.67  & 32.00  & 30.67  & 1.33  & 23.80  \\
    pGlo\tnote{*}  & 31.67  & 49.00  & 30.33  & 9.67  & 36.00  & 47.33  & 8.00  & 35.67  & 69.67  & 3.67  & 32.10   \\
    Loc-Geo & 88.00  & 96.33  & 91.33  & 100.00  & 42.67  & 57.67  & 6.33  & 97.00  & 99.67  & 1.33  & 68.03    \\
    Surf  & 66.00  & 65.67  & 63.00  & 66.67  & 50.67  & 44.00  & 28.33  & 66.67  & 62.67  & 19.00  & 53.27   \\
    Occ+Surf & 99.67  & 84.00  & 78.00  & 96.33  & 46.67  & 37.00  & 1.67  & 98.33  & 84.33  & 6.33  & 63.23   \\
    Occ+Surf+pGlo & 62.00  & 95.67  & 62.00  & 84.33  & 59.67  & 54.33  & 43.33  & 84.33  & 95.00  & 1.67  & 64.23    \\
    DexRep+pGlo & 87.00  & 98.00  & 97.33  & 100.00  & 78.67  & 70.00  & 28.67  & 100.00  & 99.67  & 2.33  & 76.17   \\
    \textbf{Ours} & 100.00  & 99.00  & 97.33  & 100.00  & 94.00  & 81.33  & 64.00  & 100.00  & 100.00  & 3.33  & \textbf{83.90}   \\
    \bottomrule
    \bottomrule
    \end{tabular}%
    
     \begin{tablenotes}
        \footnotesize
        \item[*] The features used in Hand2obj are from  DAPG~\cite{DAPG:rajeswaran2017learning} and pGlo from ILAD~\cite{ILAD:wu2022learning}. 
      \end{tablenotes}
\end{threeparttable}
  \label{tab:success rate of 10 novel objects of GRAB}%
\end{table*}%

\begin{figure*}[ht]
    \begin{minipage}{0.52\linewidth}
\includegraphics[width=\linewidth]{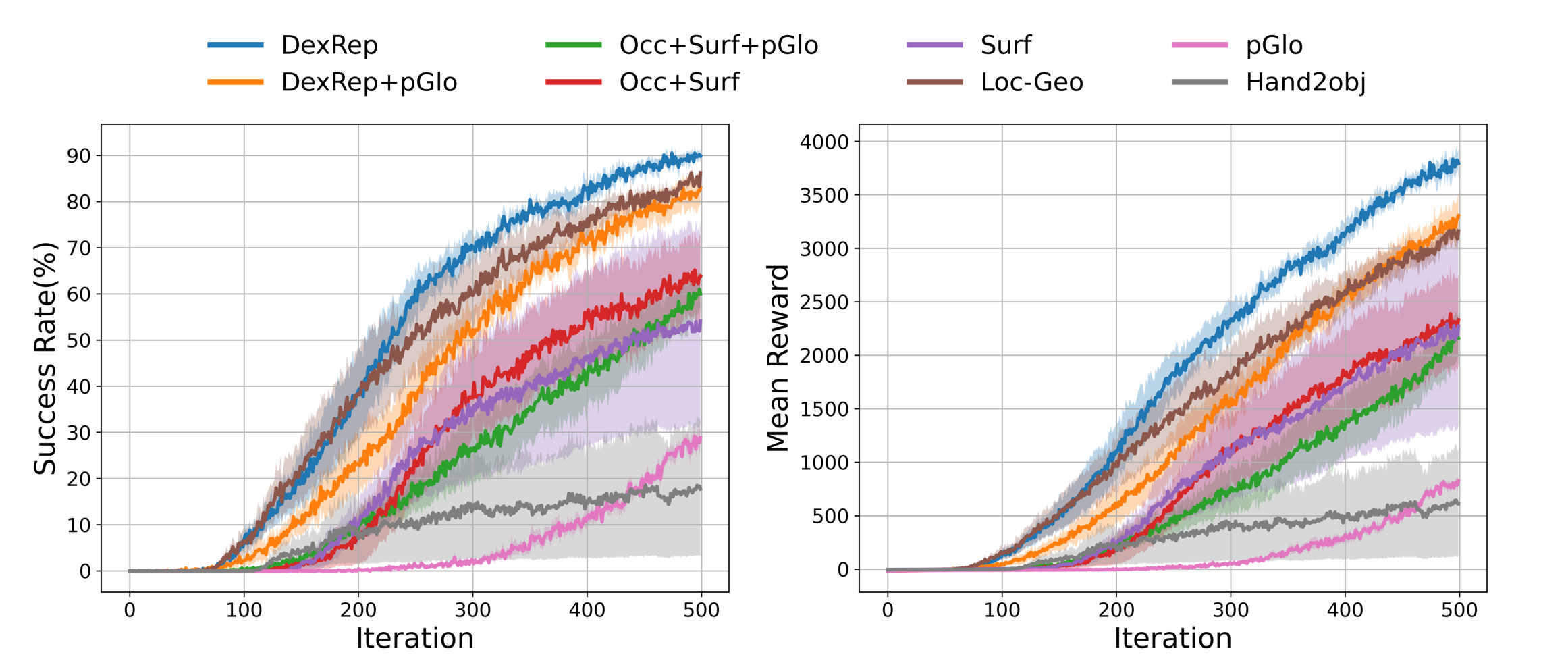}
    \caption{Mean reward and success rate of our method and baselines during RL training}  
    \label{train}
    \end{minipage}
\hfill
    \begin{minipage}{0.44\linewidth}
\includegraphics[width=\linewidth]{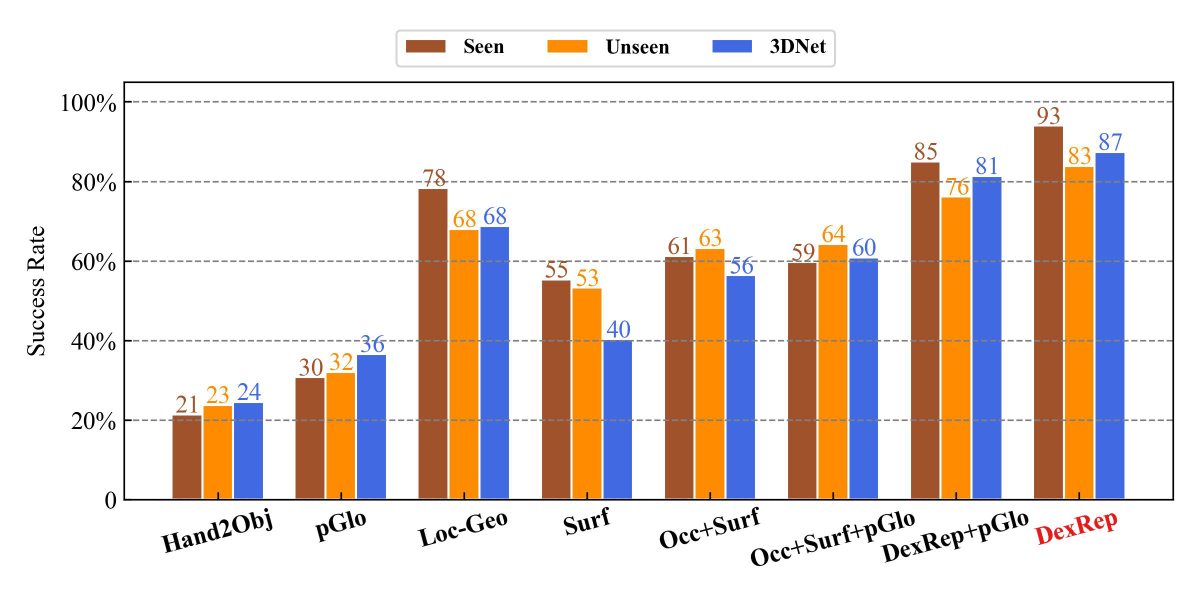}
      \caption{Qualitative results of our method and baselines on 50 objects(40 seen objects; 10 unseen objects) from GRAB and 30 unseen objects from 3DNet.}
      \label{test}
    \end{minipage}
    \vskip -0.5cm
\end{figure*}

\subsection{Experiment settings} \label{exp settings}

\textbf{Simulation environment.} We conduct simulations using the MuJoCo~\cite{todorov2012mujoco} environment, shown in Fig.\ref{fig:expinit}. At the start of each episode, the object's position is set to the origin of the world coordinate system, with a disturbance of $\left [ -0.05,0.05 \right ] $ in the x-axis direction and $\left [ -0.05,0 \right ] $ in the y-axis direction. Moreover, the object is randomly rotated around the z-axis within the range of $\left [ -\pi,\pi \right ] $. Its mass $m_0$ and friction $\mu$ are set to $0.5kg$ and $1$ respectively. We employ the 30-DoFs AdroitHand~\cite{kumar2013adroithand} model in our experiments and initialize it with the average pose of the first steps in all retargeted demonstrations, which ensures a natural pre-grasp pose and keeps the object within the effective range of the feature discussed in \secref{dexrep}. 


\textbf{Dataset.} We generate retargeted demonstrations with the GRAB dataset. We use 40 object CAD models provided by GRAB~\cite{taheri2020grab} for training and the remaining 10 objects for evaluation. In addition, we select 30 extra objects from 3DNet~\cite{wohlkinger20123dnet} that are not seen during training to validate our method. Unseen objects are shown in Fig.\ref{fig:expinit}. 

\textbf{Metrics.} We use the success rate of grasping as the evaluation metric for the quality of our method. In one episode, if an object is successfully grasped and held within $10cm$ of the target position for at least 50-time steps, we consider the grasp to be successful.

\textbf{Implementation details.} We conducted training for the grasping policy on a system equipped with Intel Xeon Gold 6326 processors and NVIDIA 3090 GPUs. During the Behavioral Cloning (BC) phase, we optimized all parameters using the Adam optimizer for 150 epochs, employing a minibatch size of 64 and a learning rate of $1e-5$. This process is typically completed in approximately 6 minutes.

In the subsequent Reinforcement Learning (RL) training phase, we used a dataset comprising 40 different objects to develop a versatile grasping policy, spanning about 500 iterations. During each iteration, we generated 10 grasping trajectories for each object, with each trajectory spanning 200 time steps (100 during evaluation). Training a generalizable grasping policy of this scale required approximately 35 GPU hours.


\subsection{Baseline} \label{baselines}
To verify the effectiveness of our proposed DexRep, we construct variants of our method. The first two variants use the representation for dexterous grasping in prior works and the others ablate the proposed features. 
\begin{enumerate}
    \item \textbf{Hand2obj:} Similar to DAPG~\cite{DAPG:rajeswaran2017learning}, this baseline only uses the relative position to describe the relationship between the hand and the object.
    \item \textbf{pGlo:}  Similar to ILAD~\cite{ILAD:wu2022learning}, this baseline uses the global features extracted by the pre-trained PointNet model and the 6D pose to train the policy.
    \item \textbf{Loc-Geo:} This baseline replace the global features in \textbf{pGlo} with our proposed Loc-Geo Feature for $n$ closest points.    
    \item \textbf{Surf:} This baseline only utilizes only our proposed Surface Feature    
    \item \textbf{Occ+Surf} This baseline only add \textbf{Occ} to \textbf{Surf}.
    \item \textbf{Occ+Surf+pGlo} We combine Occupancy Feature and Surface Feature with the global feature extracted from PointNet.    
    \item \textbf{DexRep+pGlo:} We add the global feature extracted from PointNet to our proposed DexRep.
    \item \textbf{DexRep (ours):} Our full method utilizes all three features, \ie \textbf{Occ+Surf+Loc-Geo}.
\end{enumerate}

\subsection{Effectiveness of DexRep} 
\label{results-in-simulation}

We train dynamic grasping policy with 3 random seeds and sample 10 grasping trajectories for each training object in MuJoCo~\cite{todorov2012mujoco} at each iteration. Fig.\ref{train} shows the training process of our method and baselines. Our method converges fastest and reaches the highest reward and success rate. 

The results for the group experiments of \textbf{Surf},  \textbf{Occ+Surf}, \textbf{DexRep} verify the effectiveness of each proposed feature. From the basic Surface Feature, adding Occupancy Feature and Loc-Geo Feature one by one keeps increasing the reward and success rate significantly.

Finer global geometry features may harm the learning of grasping policy. 
Comparing our method with the baseline \textbf{DexRep+pGlo}, or \textbf{Occ+Surf} with \textbf{Occ+Surf+pGlo}, we can see from Fig. \ref{train} that 
adding global features of the object extracted by PointNet does not improve the learning but makes the convergence curves lower.
On the other hand, local geometric features boost the performance (both convergence speed and reward) of grasping dramatically from \textbf{pGlo} and \textbf{Occ+Surf}, indicating that compared with global features, finer or coarse, the high-quality local feature is the key feature for dexterous grasping. The main reason for the contrast in the performance is that local geometric features can be shared across many different objects. For example, for objects having similar handles with various body shapes, our proposed local features for handles are similar and therefore, the learned policy can grasp unseen objects via handles successfully.



To validate the generalization ability of different methods to unseen objects, we evaluate the trained policy with 3 random seeds on 10 unseen objects from the GRAB dataset~\cite{taheri2020grab} and 30 unseen objects in 3DNet~\cite{wohlkinger20123dnet}. The result is shown in Fig. \ref{test}. Also, we provide detailed results of the success rates for each unseen object of GRAB in Table \ref{tab:success rate of 10 novel objects of GRAB}. Our method outperforms baselines and the success rate of our method is at least 50\% higher than the features used in prior works~\cite{DAPG:rajeswaran2017learning, ILAD:wu2022learning} after 500 training iterations, indicating that the policy trained by our method can better learn the commonalities among different objects and generalize to unseen objects.

\begin{figure}
    \centering
    \includegraphics[width=0.8\linewidth]{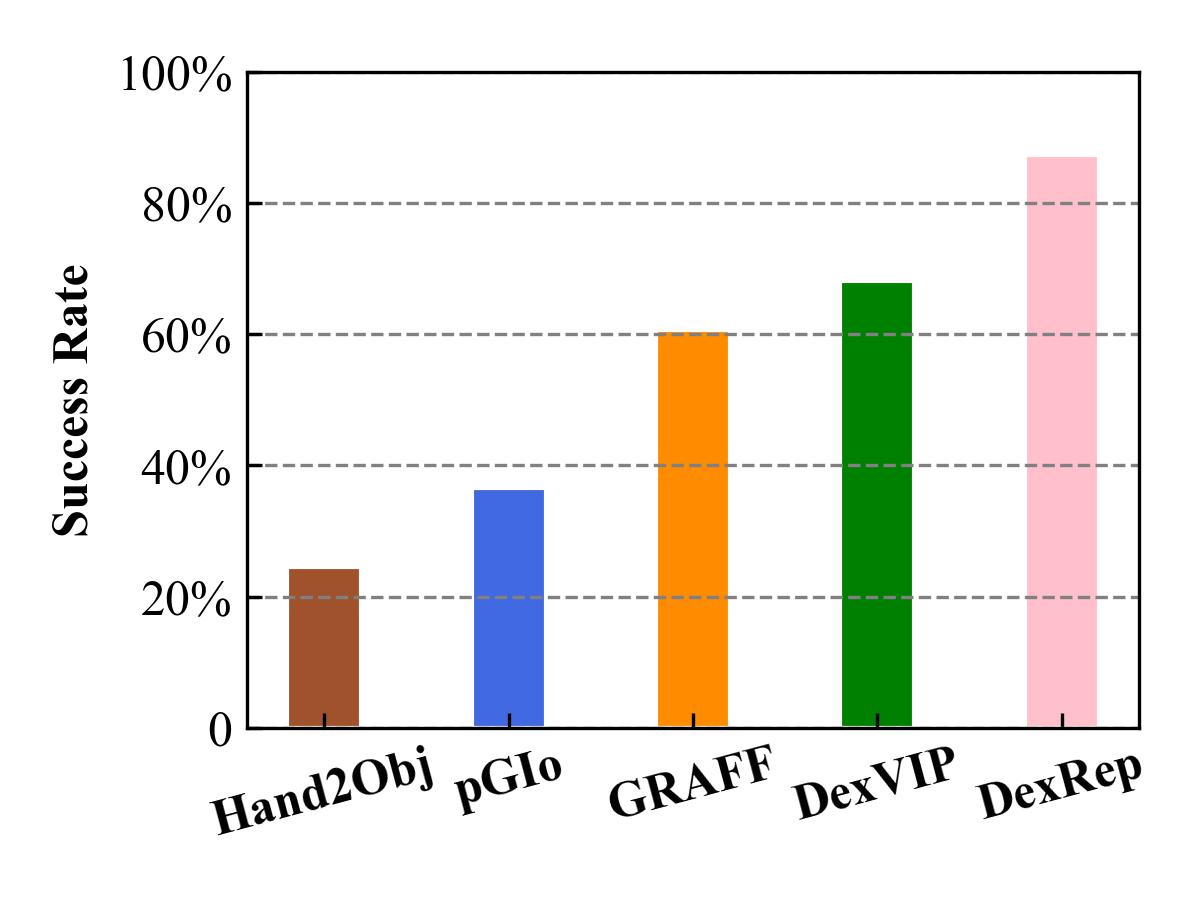}
    \caption{Qualitative results of the comparison between different methods on novel objects from 3DNet}
    \label{fig:dexvip}
    \vskip -0.5cm
\end{figure}
In addition, we compare our method with state-of-the-art RL grasping methods DexVIP~\cite{mandikal2022dexvip} and GRAFF~\cite{GRAFF:mandikal2021learning}, which use 24 unseen objects from 3DNet~\cite{wohlkinger20123dnet} to test their success rate of grasping \footnote{The objects in DexVIP and GRAFF are not detailed in their papers. For our evaluation, we randomly sample objects from categories not in the GRAB training set from the Cat60 subclasses of 3DNet. }. Together with \textbf{pGlo} and \textbf{Hand2obj}, the compared results are shown in Fig.\ref{fig:dexvip}.

\subsection{Application to Multi-morphology Robotic Hands} \label{Multi-mophology robotic hands} 

\begin{table}[htbp]
 \centering
   \tabcolsep=11.8pt
   \small
  \caption{Success rate (\%) of policies on unseen objects with multi-finger robotic hands}
    \begin{tabular}{c|c|c|c}
    \toprule
    \toprule
    num of finger & Hand2obj & pGlo & \textbf{Ours} \\
    \midrule
    2     & 8.97  & 38.27 & \textbf{68.77} \\
    3     & 17.90  & 56.40  & \textbf{81.80} \\
    4     & 0.00     & 18.60  & \textbf{83.90} \\
    \bottomrule
    \bottomrule
    \end{tabular}%
  \label{tab:mulhand}%
\end{table}%

\begin{figure}[htbp]
    \centering
    \includegraphics[trim=0cm 0cm 0cm 0cm, clip, width=0.8\linewidth]{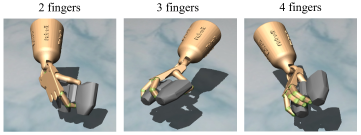}
    \caption{Examples of grasping unseen objects with different robotic hands varying numbers of fingers.}
    \label{fig:diffrent morphogries}
    \vskip -0.5cm
\end{figure}

The experiment studies the application of \rep on robotic variant hands. We obtain multiple dexterous robotic hands with various configurations by disassembling the fingers of the five-fingered hand as \cite{Dissemble：radosavovic2021state-only} does, as shown in Fig.\ref{fig:diffrent morphogries}. We compare our method with \textbf{pGlo} and \textbf{Hand2obj}, whose results are shown in Table \ref{tab:mulhand}. In the task of robotic dexterous grasping, our proposed approach is applicable to multiple types of multi-joint robotic hands and maintains good grasping performance and generalization ability. 


\subsection{Real experiment} \label{results-in-the-real-world}
\begin{figure}[htbp]
    \centering
    \includegraphics[width=0.9\linewidth]{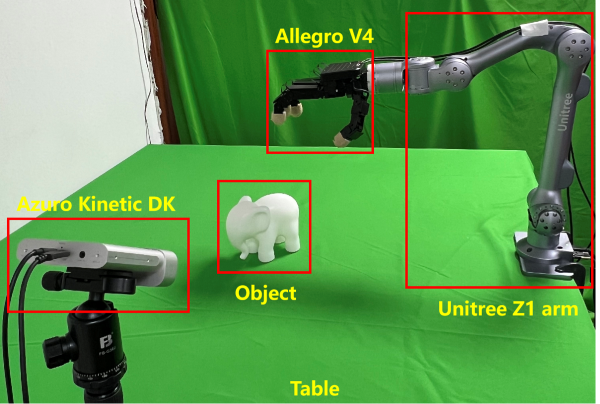}
    \caption{The grasping platform with an Allegro Hand v4,  a Unitree Z1 arm and an Azure Kinetic DK.}
    \label{fig:hardware system}
\end{figure}

\begin{figure}[htbp]
     \centering
     \includegraphics[width=0.48\textwidth]{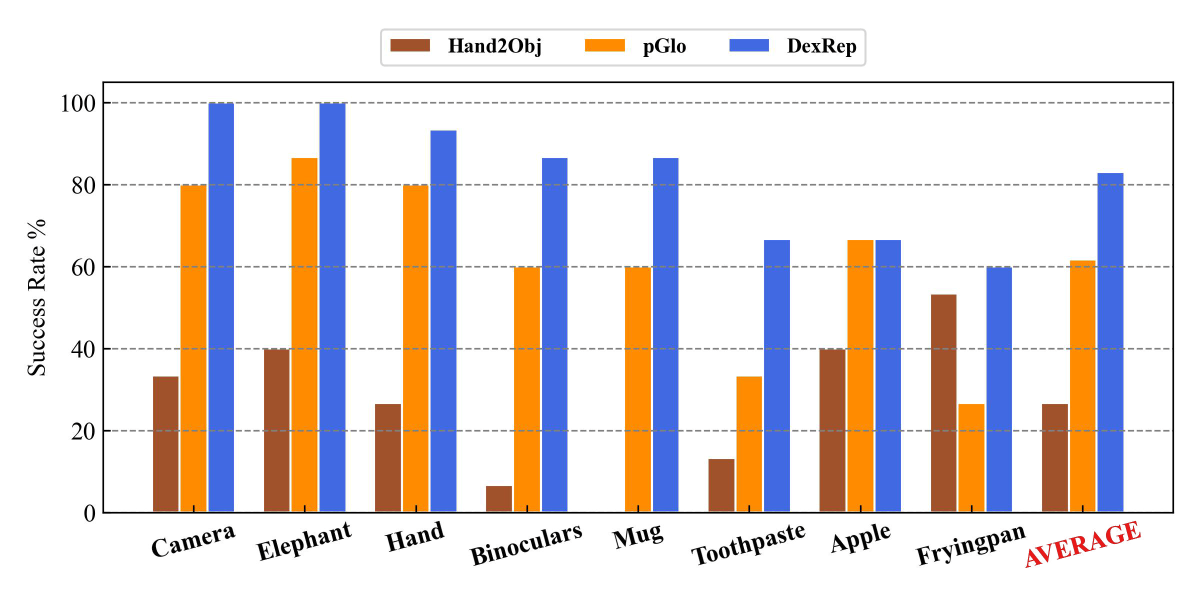}
     \caption{Results of our real-world experiment. }
     \label{fig:RealExp SuccRate}
     \vskip -0.5cm
 \end{figure}

To validate the effectiveness of the trained policy, we build up a grasp system consisting of an Allegro Hand~\footnote{https://www.wonikrobotics.com/research-robot-hand}, a Unitree Z1 arm~\footnote{https://www.unitree.com/arm/}, and an Azure Kinect DK~\footnote{https://azure.microsoft.com/en-us/products/kinect-dk}, which is shown in Fig.\ref{fig:hardware system}. In our validation system, we assume object CAD models are known. Therefore, only 6D poses of the objects are required to grasp,  which are obtained by registering the CAD models to the partial point clouds captured from the depth sensor of Azure Kinect DK. The objects are segmented out from green background from RGB images calibrated with the depth sensor of Azure Kinect DK and then the segmented depth images are back-projected back into 3D space to get partial point clouds. To mitigate the influence of the pose registration errors, we added Gaussian noise with variances of $2cm$ for position and $0.1rad$ for rotation to the object poses during the policy training. 



We evaluate our method and two baselines on 8 unseen objects in the real experiment. Each object was grasped 15 times to calculate its success rate. As shown in Fig.\ref{fig:RealExp SuccRate}, our method outperforms the baselines and demonstrates excellent generalization capability on unseen objects even in the real world.

%% file: text/conclusion.tex
\section{conclusion}


In this paper, we proposed a novel hand-object interaction representation for robotic dexterous grasping, called \rep, which consists of Occupancy Feature, Surface Feature, and Local-Geo Feature. This representation captures the relative shape feature of the objects and the spatial relation between hands and objects during hand-object interactions, enabling generalization to novel objects. Based on this representation, we proposed a dexterous deep reinforcement learning method, named \method, to train a generalizable grasping policy. Experimental results demonstrate the effectiveness of our proposed representation and method. Compared to baselines, our method achieves higher grasp success rates on both seen and unseen objects and also shows better generalization ability in both simulation and the real world. In summary, our proposed method provides a promising solution for robotic dexterous grasping in unstructured environments, enabling robots to perform complex tasks and achieve human-like capabilities. 


\section{acknowledgments}

This work was supported in part by NSFC under Grants 62088101, 62103372, and 62233013.

%% file: root.bbl
\begin{thebibliography}{10}
\providecommand{\url}[1]{#1}
\csname url@rmstyle\endcsname
\providecommand{\newblock}{\relax}
\providecommand{\bibinfo}[2]{#2}
\providecommand\BIBentrySTDinterwordspacing{\spaceskip=0pt\relax}
\providecommand\BIBentryALTinterwordstretchfactor{4}
\providecommand\BIBentryALTinterwordspacing{\spaceskip=\fontdimen2\font plus
\BIBentryALTinterwordstretchfactor\fontdimen3\font minus
  \fontdimen4\font\relax}
\providecommand\BIBforeignlanguage[2]{{%
\expandafter\ifx\csname l@#1\endcsname\relax
\typeout{** WARNING: IEEEtran.bst: No hyphenation pattern has been}%
\typeout{** loaded for the language `#1'. Using the pattern for}%
\typeout{** the default language instead.}%
\else
\language=\csname l@#1\endcsname
\fi
#2}}

\bibitem{DAPG:rajeswaran2017learning}
A.~Rajeswaran, V.~Kumar, A.~Gupta, G.~Vezzani, J.~Schulman, E.~Todorov, and
  S.~Levine, ``{Learning complex dexterous manipulation with deep reinforcement
  learning and demonstrations},'' in \emph{RSS}, 2018.

\bibitem{ILAD:wu2022learning}
Y.-H. Wu, J.~Wang, and X.~Wang, ``Learning generalizable dexterous manipulation
  from human grasp affordance,'' in \emph{CoRL}.\hskip 1em plus 0.5em minus
  0.4em\relax PMLR, 2023, pp. 618--629.

\bibitem{GRAFF:mandikal2021learning}
P.~Mandikal and K.~Grauman, ``Learning dexterous grasping with object-centric
  visual affordances,'' in \emph{ICRA}, 2021.

\bibitem{mandikal2022dexvip}
{P. Mandikal and K. Grauman}, ``Dexvip: Learning dexterous grasping with human
  hand pose priors from video,'' in \emph{CoRL}, 2022.

\bibitem{qi2017pointnet}
C.~R. Qi, H.~Su, K.~Mo, and L.~J. Guibas, ``Pointnet: Deep learning on point
  sets for 3d classification and segmentation,'' in \emph{CVPR}, 2017.

\bibitem{wei2022dvgg}
W.~Wei, D.~Li, P.~Wang, Y.~Li, W.~Li, Y.~Luo, and J.~Zhong, ``Dvgg: Deep
  variational grasp generation for dextrous manipulation,'' \emph{RAL}, 2022.

\bibitem{shao2020unigrasp}
L.~Shao, F.~Ferreira, M.~Jorda, V.~Nambiar, J.~Luo, E.~Solowjow, J.~A. Ojea,
  O.~Khatib, and J.~Bohg, ``Unigrasp: Learning a unified model to grasp with
  multifingered robotic hands,'' \emph{RAL}, 2020.

\bibitem{ye2022CGF}
J.~Ye, J.~Wang, B.~Huang, Y.~Qin, and X.~Wang, ``Learning continuous grasping
  function with a dexterous hand from human demonstrations,'' \emph{RAL},
  vol.~8, no.~5, pp. 2882--2889, 2023.

\bibitem{He_2016_CVPR}
K.~He, X.~Zhang, S.~Ren, and J.~Sun, ``Deep residual learning for image
  recognition,'' in \emph{CVPR}, 2016.

\bibitem{levine2018handeye}
S.~Levine, P.~Pastor, A.~Krizhevsky, J.~Ibarz, and D.~Quillen, ``Learning
  hand-eye coordination for robotic grasping with deep learning and large-scale
  data collection,'' \emph{The International Journal of Robotics Research},
  2018.

\bibitem{taheri2020grab}
O.~Taheri, N.~Ghorbani, M.~J. Black, and D.~Tzionas, ``Grab: A dataset of
  whole-body human grasping of objects,'' in \emph{ECCV}, 2020.

\bibitem{handa2020dexpilot}
A.~Handa, K.~Van~Wyk, W.~Yang, J.~Liang, Y.-W. Chao, Q.~Wan, S.~Birchfield,
  N.~Ratliff, and D.~Fox, ``Dexpilot: Vision-based teleoperation of dexterous
  robotic hand-arm system,'' in \emph{ICRA}, 2020.

\bibitem{liang2021multifingered}
H.~Liang, L.~Cong, N.~Hendrich, S.~Li, F.~Sun, and J.~Zhang, ``Multifingered
  grasping based on multimodal reinforcement learning,'' \emph{RAL}, 2021.

\bibitem{turpin2022graspd}
D.~Turpin, L.~Wang, E.~Heiden, Y.-C. Chen, M.~Macklin, S.~Tsogkas,
  S.~Dickinson, and A.~Garg, ``Grasp’d: Differentiable contact-rich grasp
  synthesis for multi-fingered hands,'' in \emph{ECCV}, 2022.

\bibitem{liu2020deep}
M.~Liu, Z.~Pan, K.~Xu, K.~Ganguly, and D.~Manocha, ``Deep differentiable grasp
  planner for high-dof grippers,'' \emph{arXiv preprint arXiv:2002.01530},
  2020.

\bibitem{Christen_2022_CVPR}
S.~Christen, M.~Kocabas, E.~Aksan, J.~Hwangbo, J.~Song, and O.~Hilliges,
  ``D-grasp: Physically plausible dynamic grasp synthesis for hand-object
  interactions,'' in \emph{CVPR}, 2022.

\bibitem{li2022contact2grasp}
H.~Li, X.~Lin, Y.~Zhou, X.~Li, J.~Chen, and Q.~Ye, ``Contact2grasp: 3d grasp
  synthesis via hand-object contact constraint,'' \emph{arXiv preprint
  arXiv:2210.09245}, 2022.

\bibitem{varley2017shape}
J.~Varley, C.~DeChant, A.~Richardson, J.~Ruales, and P.~Allen, ``Shape
  completion enabled robotic grasping,'' in \emph{IROS}, 2017.

\bibitem{cao2021suctionnet}
H.~Cao, H.-S. Fang, W.~Liu, and C.~Lu, ``Suctionnet-1billion: A large-scale
  benchmark for suction grasping,'' \emph{RAL}, 2021.

\bibitem{yuan2017bighand2}
S.~Yuan, Q.~Ye, B.~Stenger, S.~Jain, and T.-K. Kim, ``Bighand2. 2m benchmark:
  Hand pose dataset and state of the art analysis,'' in \emph{CVPR}, 2017, pp.
  4866--4874.

\bibitem{joshi2020robotic}
S.~Joshi, S.~Kumra, and F.~Sahin, ``Robotic grasping using deep reinforcement
  learning,'' in \emph{CASE}, 2020.

\bibitem{zhang2021manipnet}
H.~Zhang, Y.~Ye, T.~Shiratori, and T.~Komura, ``Manipnet: neural manipulation
  synthesis with a hand-object spatial representation,'' \emph{ACM ToG}, 2021.

\bibitem{kumar2013adroithand}
V.~Kumar, Z.~Xu, and E.~Todorov, ``Fast, strong and compliant pneumatic
  actuation for dexterous tendon-driven hands,'' in \emph{ICRA}.\hskip 1em plus
  0.5em minus 0.4em\relax IEEE, 2013, pp. 1512--1519.

\bibitem{chang2015shapenet}
A.~X. Chang, T.~Funkhouser, L.~Guibas, P.~Hanrahan, Q.~Huang, Z.~Li,
  S.~Savarese, M.~Savva, S.~Song, H.~Su, \emph{et~al.}, ``Shapenet: An
  information-rich 3d model repository,'' \emph{arXiv preprint
  arXiv:1512.03012}, 2015.

\bibitem{wohlkinger20123dnet}
W.~Wohlkinger, A.~Aldoma, R.~B. Rusu, and M.~Vincze, ``3dnet: Large-scale
  object class recognition from cad models,'' in \emph{ICRA}, 2012.

\bibitem{MANO:SIGGRAPHASIA:2017}
J.~Romero, D.~Tzionas, and M.~J. Black, ``Embodied hands: Modeling and
  capturing hands and bodies together,'' \emph{ACM Transactions on Graphics,
  (Proc. SIGGRAPH Asia)}, vol.~36, no.~6, Nov. 2017.

\bibitem{todorov2012mujoco}
E.~Todorov, T.~Erez, and Y.~Tassa, ``Mujoco: A physics engine for model-based
  control,'' in \emph{IROS}.\hskip 1em plus 0.5em minus 0.4em\relax IEEE, 2012,
  pp. 5026--5033.

\bibitem{chen2022dextransfer}
Z.~Q. Chen, K.~Van~Wyk, Y.-W. Chao, W.~Yang, A.~Mousavian, A.~Gupta, and
  D.~Fox, ``Dextransfer: Real world multi-fingered dexterous grasping with
  minimal human demonstrations,'' \emph{arXiv preprint arXiv:2209.14284}, 2022.

\bibitem{GAE:schulman2015high}
J.~Schulman, P.~Moritz, S.~Levine, M.~Jordan, and P.~Abbeel, ``High-dimensional
  continuous control using generalized advantage estimation,'' \emph{arXiv
  preprint arXiv:1506.02438}, 2015.

\bibitem{Dissemble：radosavovic2021state-only}
I.~Radosavovic, X.~Wang, L.~Pinto, and J.~Malik, ``State-only imitation
  learning for dexterous manipulation,'' in \emph{IROS}, 2021.

\end{thebibliography}
